\documentclass[10pt,twocolumn,letterpaper]{article}%

\usepackage{cvpr}
\usepackage{times}
\usepackage{epsfig}
\usepackage{graphicx}
\usepackage{amsmath}
\usepackage{amssymb}

\usepackage{color}
\definecolor{green}{rgb}{0, 0.5, 0}
\definecolor{orange}{rgb}{0.8, 0.6, 0.2}
\definecolor{red}{rgb}{1.0, 0.0, 0.0}
\definecolor{teal}{rgb}{0.0, 0.4, 0.4}
\definecolor{purple}{rgb}{0.65,0,0.65}
\definecolor{saffron}{rgb}{0.95,0.75,0.2}
\definecolor{turquoise}{rgb}{0.0,0.5,0.5}

\usepackage[ruled,vlined,linesnumbered]{algorithm2e}
\usepackage{multirow}

\newcommand{\kx}[1]{{\color{black}#1}}

\newcommand{\wxg}[1]{{\color{black}#1}}

\usepackage[normalem]{ulem}

\usepackage{mathtools}
\usepackage{amsmath, amssymb, amsfonts, amsopn}  %amsthm,
\usepackage{graphicx} % Necessary to use \scalebox
\usepackage{textcomp}
\usepackage{xfrac}
\usepackage{bbm}

\usepackage{overpic}
\usepackage{subfig}
\usepackage{wrapfig}

\newcommand{\hidecomment}[1]{}
\usepackage[pagebackref=true,breaklinks=true,letterpaper=true,colorlinks,bookmarks=false]{hyperref}

\cvprfinalcopy % *** Uncomment this line for the final submission

\def\cvprPaperID{5312} % *** Enter the CVPR Paper ID here

\ifcvprfinal\fi

\begin{document}

%%%%%%%%% TITLE
\title{Learning Fine-Grained Segmentation of 3D Shapes without Part Labels}

\author{
Xiaogang Wang$^{1,2}$ \quad Xun Sun$^2$ \quad Xinyu Cao$^2$ \quad Kai Xu$^{3*}$ 
\quad Bin Zhou$^2$\thanks{Corresponding author: kevin.kai.xu@gmail.com,zhoubin@buaa.edu.cn}\\
$^1$Southwest University\\
$^2$State Key Laboratory of Virtual Reality Technology and Systems, Beihang University\\
$^3$National University of Defense Technology\\
}

\maketitle
\thispagestyle{empty}

\begin{abstract}
%Fine-grained segmentation of 3D shapes is an important task for 3D vision, many
Learning-based 3D shape segmentation is usually formulated
as a semantic labeling problem, assuming that all parts of training shapes are annotated
with a given set of tags. This assumption, however, is impractical for learning fine-grained
segmentation. Although most off-the-shelf CAD models are, by construction, composed of fine-grained parts,
they usually miss semantic tags and labeling those fine-grained parts is extremely tedious.
We approach the problem with deep clustering, where
the key idea is to learn part priors from a shape dataset with fine-grained segmentation
but no part labels.
Given point sampled 3D shapes, we model the clustering priors of points with a similarity matrix
and achieve part segmentation through minimizing a novel low rank loss.
%Further, since fine-grained parts can be very tiny,
%a 3D shape has to be densely sampled to ensure the tiny parts are well captured and segmented.
To handle highly densely sampled point sets, we adopt a divide-and-conquer strategy.
We partition the large point set into a number of blocks.
Each block is segmented using a deep-clustering-based part prior network trained in a category-agnostic manner.
We then train a graph convolution network to merge the segments of all blocks
to form the final segmentation result.
Our method is evaluated with a challenging benchmark of fine-grained segmentation, showing state-of-the-art performance.
%For the segmentation of point sets, we can actually consider that the points features within the same
%part have stronger similarity, at the same time, the features of points between different parts should be as dissimilar as possible. According to this idea, we defined a similarity matrix to describe this phenomenon.
%2), According to the low rank of the similarity matrix, we designed a novel low rank loss, which can effectively perform low-rank approximation of the similarity matrix and solve the problem of unequal number of parts in each model.
%The low-rank matrix is aslo the segmentation result of the model.
%Meanwhile, in order to guarantee enough point sampling over small fine-grained parts in 3D shape, we need to sample more points for each 3D shape. Existing point cloud-based methods are expensive to directly deal with large point sets. Because they require a huge amount of GPU memory and computation. To this end, we design a divide-and-conquer strategy: the first stage, large point sets is partitioned as some blocks. we develop a neural network Patch-Net for each block segmentation. The second stage, we design a graph convolutions-based network, Merge-Net, to merge the segmentation results of adjacent blocks to obtain the fine-grained segmentation results. Our method is evaluated with an extremely challenging benchmark for fine-grained analysis of 3D shapes. Results demonstrate that our method achieves the state-of-the-art performance.
%\keywords{3D shape segmentation, fine-grained segmentation, deep clustering}
\end{abstract}

\section{Introduction}
%Motivation

3D shape segmentation is a fundamental problem in 3D vision.
While most existing works focus on semantic segmentation of 3D shapes into major parts (e.g., seat, back and leg of a chair), many application scenarios, on the other hand, demand fine-grained shape segmentation.
\wxg{A definition of fine-grained parts was given in ~\cite{wang2018learning}. In that
work, fine-grained parts are defined in contrast with semantic parts. 
While semantic parts are major, functional ones (e.g., the back, seat and leg parts of a chair), fine-grained
parts mainly refer to modeling components which are, albeit geometrically insignificant, conceptually meaningful in the
sense of assembly-based 3D modeling ~\cite{zhu2018scores}. 
Therefore,} fine-grained segmentation induces an intricate structural analysis of 3D objects, which facilitates
part-based shape synthesis and modeling~\cite{mitra2013structure,xu2016data} and
meticulous robotic manipulation~\cite{aleotti20123d,hu2016}.
Consequently, the problem receives increasing research attention lately~\cite{yu2019partnet,wang_siga18}, along with dedicated datasets~\cite{mo2018partnet}.

\begin{figure}[t!]\centering
  \begin{overpic}[width=1.0\linewidth,tics=10]{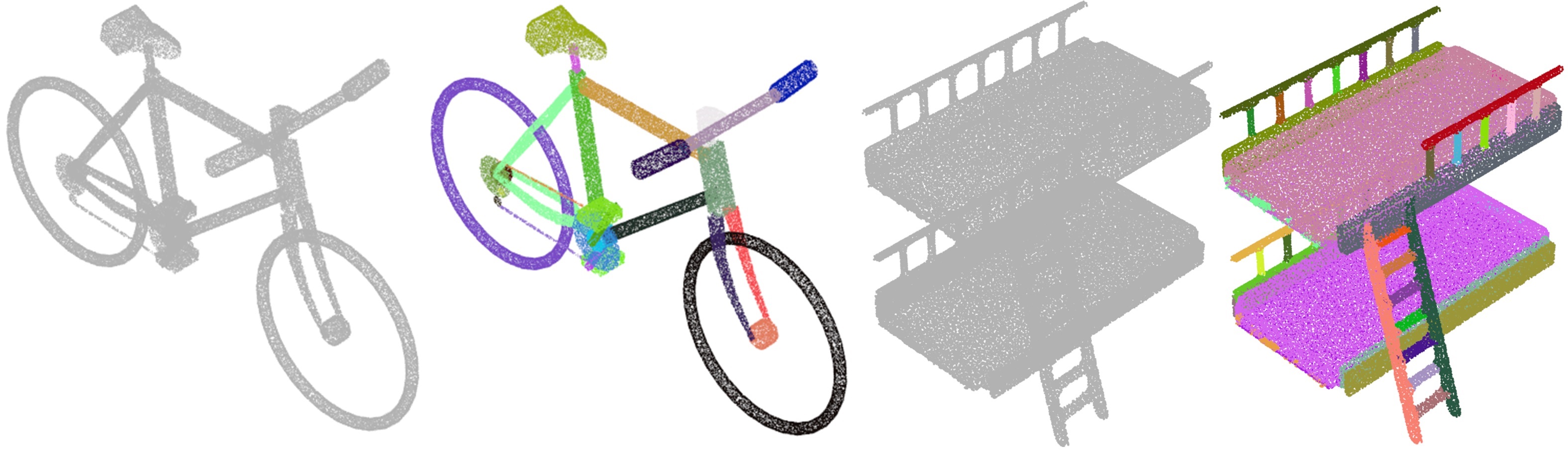}%,grid
  %\put(30,29){\small (a)}
  %\put(78,29){\small (b)}
  \put(30,-2){\small (a)}
  \put(78,-2){\small (b)}
  \end{overpic}
  \caption{Two examples of fine-grained segmentation. For each example, the left is the input point cloud and the right is the fine-grained segmentation result.}
  \label{fig:teaser}\vspace{-8pt}
\end{figure}

Previous learning-based approaches to fine-grained 3D shape segmentation usually formulate it
as a semantic labeling problem. This requires a large training dataset of 3D shapes with fine-grained part segmentation and tags.
When working with most online shape repositories such as ShapeNet~\cite{ShapeNet2015}, fine-grained part segmentation comes for free
since most off-the-shelf CAD models are, by construction, composed of fine-grained parts.
These fine-grained parts, however, have no, or noisy semantic tags.
Annotating fine-grained parts with semantic tags is extremely tedious due to the tiny part size and large part count (range from tens to hundreds; see~\cite{wang_siga18} for statistics). Moreover, many fine-grained parts may not even have a well-defined tag.
Due to these reasons, the existing fine-grained part datasets~\cite{mo2018partnet} does leave many parts unlabeled.

In this paper, we introduce a \emph{deep clustering} based approach to fine-grained part segmentation, thus avoiding the requirement of part labels.
The key idea is to learn \emph{geometric part priors} describing \emph{what constitutes a fine-grained part},
based on a shape dataset with fine-grained segmentation but no part labels.
Working with point sampled 3D shapes,
our method models the clustering priors of 3D points with a similarity matrix of point features
capturing for any two points how likely they belong to the same part.
This similarity matrix possesses low rank property with the rank equal to the number of fine-grained parts of the shape.
Therefore, fine-grained part segmentation can be achieved by minimizing a novel low rank loss over the similarity matrix.
%The resulting low-rank matrix indicates the final segmentation.

\begin{figure}[t]
  \centering
  \includegraphics[width=\linewidth]{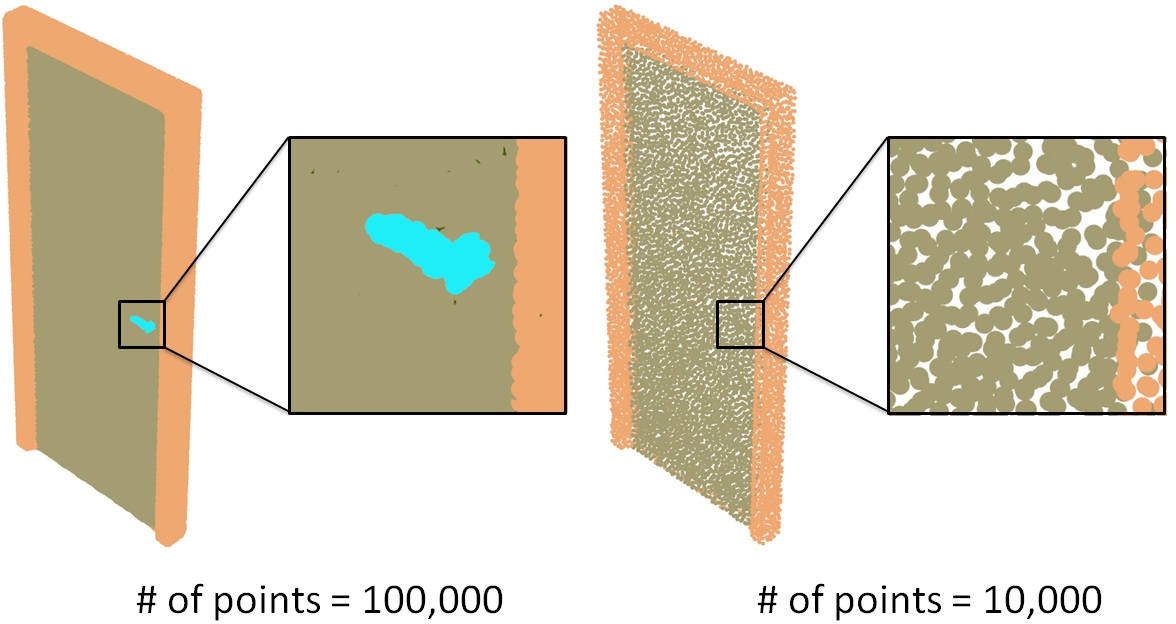}
  \caption{\kx{Tiny parts demand high sampling rate for accurate fine-grained segmentation.}}
  \label{fig:tiny}
\end{figure}

Fine-grained parts are usually very tiny compared to the full shape (see Figure~\ref{fig:tiny} (a)).
A moderate sampling rate of 3D shapes can hardly capture the
geometry of tiny parts accurately, which can result in suboptimal segmentation (see Figure~\ref{fig:tiny} (b)).
Therefore, fine-grained segmentation needs to work with densely sampled shapes.
Existing deep learning models for 3D point clouds, such as PointNet~\cite{qi2016pointnet},
usually find difficulty in handling very large point clouds.
To this end, we adopt a divide-and-conquer strategy.
We first partition the large point set into a number of blocks.
Each block is segmented using a deep-clustering-based part prior network, called PriorNet,
which is trained in a category-agnostic manner.
Benefiting from the block-wise training strategy, the required training shapes are greatly reduced.
We then train MergeNet, a graph convolution network, to merge the segments of all blocks
to form the final segmentation.

The main contributions of our paper include:
\begin{itemize}
  \vspace{-5pt}\item a deep-clustering-based formulation for fine-grained segmentation of 3D shapes which learns geometric part priors without relying on part annotations,
  \vspace{-5pt}\item a novel low-rank loss designed for learning fine-grained part priors, and
  \vspace{-5pt}\item a novel graph convolution network based module trained to merge segments in different blocks.
\end{itemize}

\begin{figure*}[t]
  \centering
  \includegraphics[width=\textwidth]{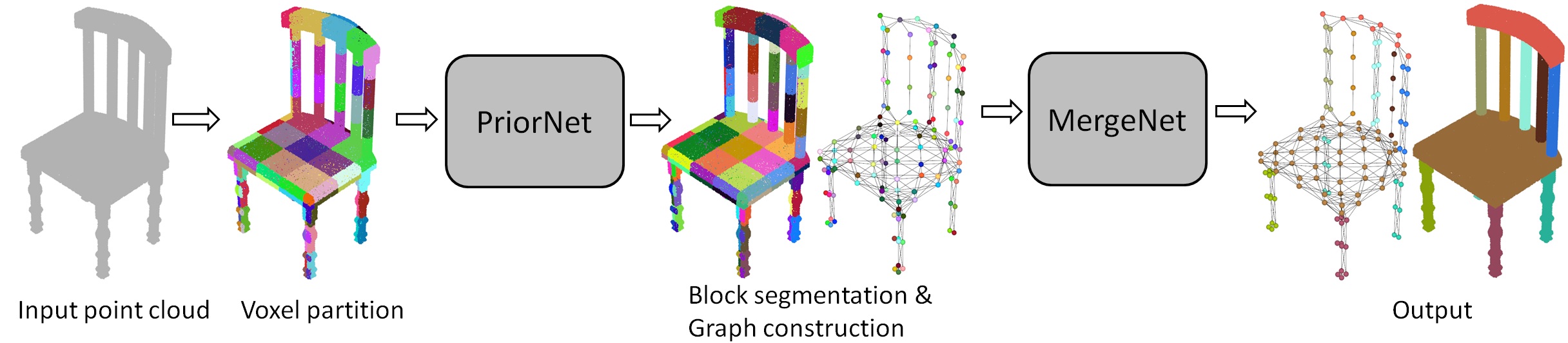}
  \caption{Pipeline overview.}
  \label{fig:pipeline}\vspace{-8pt}
\end{figure*}

\section{Related Work}

\paragraph{Point cloud segmentation.}
Point cloud segmentation has gained significant research progress in recent year, benefiting from the advances in machine learning techniques ~\cite{Su_CVPR17,qi2017pointnetplusplus,NIPS2018_7362,wang2018sgpn}.
Early studies ~\cite{rusu2009fast,rusu2008aligning,ling2007shape,chen2003visual}
most utilize hand-crafted features towards specific tasks. These features often encode statistical properties of points and are designed to be invariant transformations, which can be categorized as local features
~\cite{bronstein2010scale,sun2009concise} and global
features~\cite{rusu2008aligning,ling2007shape,johnson1999using,chen2003visual}.
For a specific task, it is not trivial to find the optimal feature combination.

Recently many deep learning architectures have been developed for point cloud data
~\cite{Su_CVPR17,qi2017pointnetplusplus,NIPS2018_7362}. These methods demonstrate remarkable performance in part segmentation of object and scene segmentation.
%The main idea is to replace the hand-crafted features employed in the traditional methods with data-driven learned ones.
All these models, however, find difficult in handling large point cloud.
In general, the point sets is down-sampled at first, which is used as the input as the input of the neural network. However,  after down-sampling, many fine-grained parts are lost.
To our knowledge, very few works have studied fine-grained segmentation of point clouds. Mo et al.~\cite{mo2018partnet} collected a large-scale dataset  with manually annotated fine-grained semantic parts.
They also proposed some baseline methods for fine-grained segmentation of 3D point cloud.
Luo et al.~\cite{luo2020learning} introduced a data-driven iterative perceptual grouping pipeline for the task of zero-shot 3D shape part discovery.
Yu et al.~\cite{yu2019partnet} proposed a top-down recursive decomposition network for fine-grained segmentation of 3D point cloud. However, their methods require well-defined fine-grained part semantic lables.

\paragraph{\textbf{Low rank representation and loss.}}
Low rank representation is a robust and efficient tool for processing high-dimensional data.
This is because low rank representation has an excellent performance in discovering global structures of data.
The low-rank representation can reveal the relationships of the samples: the within-cluster affinities are dense while the  between-cluster affinities are all zeros ~\cite{6180173}.
Low rank representation has been widely used in many applications of image processing including image denoising
~\cite{NIPS2009_3704}, face recognition ~\cite{ChenWW12a}, and classification ~\cite{ZhangLTXLM11} in recent years.

Recently, Yi et al.~\cite{yi2018faces} introduced a low rank loss based on deep learning for estimating detailed scene illumination using human faces in a single image.
The strategy based on the observation that the diffuse chromaticity over a face should be consistent among images,
regardless of illumination changes, because a person's facial surface features should remain the same.
The diffuse chromaticity of multiple aligned images of the same face should form a low rank matrix (ideally rank one), so they define the low rank loss based on the second singular value.
\wxg{Similarly, Zhu et al.~\cite{zhu2020adacoseg} proposed a low-rank loss for 3D shape co-segmentation.
The low-rank loss in AdaCoSeg measures the geometric similarity of the same semantic part across different shapes. 
The network is trained to minimize the rank; no actual low-rank decomposition is conducted.}
\wxg{Our low-rank loss is fundamentally different from the one used in ~\cite{yi2018faces,zhu2020adacoseg}.
our network exploits the low-rank and symmetric property of similarity matrix for point clustering, 
it learns to perform symmetric low-rank decomposition which directly leads to point cloud segmentation.}
To the best of our knowledge, our work is the first that defines a \wxg{loss based on low-rank decomposition} in training a point cloud segmentation network.

\paragraph{\textbf{Graph Neural Networks (GNNs).}}
Over past a few years, a series of graph-based optimization and subsequent variants achieved promising results in various applications ~\cite{zhao2018triangle, zhou2018graph}.
%including but not limited social networks ~\cite{chen2018fastgcn}, computer vision ~\cite{wang2018zero}.
%and natural language processing ~\cite{yao2019graph}.
%A more comprehensive survey on graph neural networks can be found in.
Landrieu et al.~\cite{landrieu2018large} propose a graph convolutional network-based framework to tackle the semantic segmentation of large-scale point clouds.
\if 0
They first partition the input point cloud into some parts of simple shape with an unsupervised way based on graph-cut.
Then, a graph-based classification framework is used for semantic segmentation.
\fi
This work is partly related to us, however, there are some important differences:
1), like other semantic segmentation frameworks, this approach~\cite{landrieu2018large} is classification-based and difficult to extend to our task.
2), our method learns part priors directly from a dataset through minimizing a novel low rank loss, not based on hand-crafted features.
3), our method assigns nodes based on clustering. We have no a softmax layer as the output layer for classification.

\if 0
Graph network takes graph structure and node attribute information as input.
According to different analysis tasks, it mainly includes the following three outputs:
node-level, edge-level, and graph-level.

Our approach is closely related to node-level outputs approaches.
they learn a node's high-level representation  by propagating  neighbor information in
an iterative manner until a stable fixed point is reached.
However, there is an important different.
Our MergeNet assigns nodes based on clustering.
We have no a softmax layer as the output layer for classification.
\fi

\section{Our Approach}
\label{sec:method}

%\paragraph{Method overview.}
Figure~\ref{fig:pipeline} gives an overview of our method pipeline.
Given a 3D shape represented by densely sampled point cloud, we first perform a volumetric partition to split
the point cloud into a number of blocks.
Each block is segmented with the PriorNet (Section~\ref{sec:PatchNet}).
The segments of all blocks are then merged with MergeNet to form the final segmentation (Section~\ref{sec:MergeNet}).
%It contains two major modules, a part prior network (PriorNet) and a segment merging network (MergeNet).

\if 0
Given a 3D point cloud, we firstly divide the point sets into blocks through voxel partition
(Section ~\ref{sec:PatchNet}). Each block is then segmented by Patch-Net (Section~\ref{sec:PatchNet}). Then, for each model, we search out the adjacent segments in the adjacent blocks, and form the candidate merge pairs (Section~\ref{sec:MergeNet}). Subsequently, each candidate merge pair is fed into a Merger-Net which predicts the match score that they belong to the same part (Section~\ref{sec:MergeNet}\ref).
Finally, a merge graph is constructed, and each node in the graph refers to the predicted segment from Patch-Net, and the edge refers to the matching score from Merge-Net between the two segments.
According to the merge graph, we design a bottom-up hierarchical aggregation algorithm to obtain the final fine-grained parsing result (Section~\ref{sec:labelling}).
In the next, we describe the three algorithmic components, including Patch-Net, Merge-Net, and hierarchical aggregation algorithm for fine-grained parsing.
\fi

\subsection{Part Prior Network (PriorNet)}
\label{sec:PatchNet}

\begin{figure*}[t]
  \centering
  \includegraphics[width=0.95\textwidth]{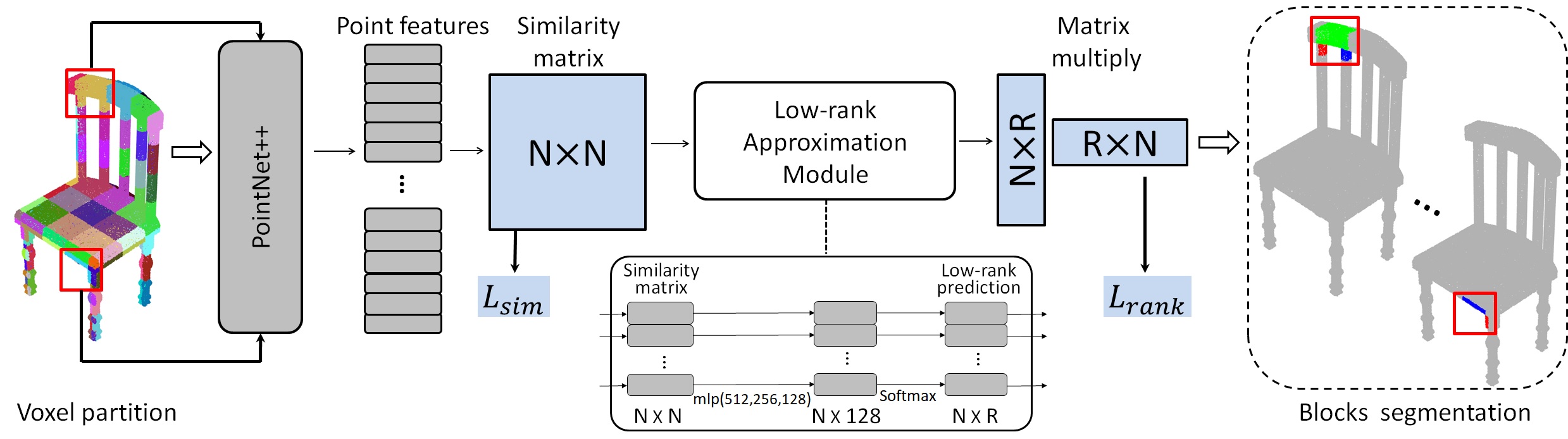}
  \caption{Network architecture of the PriorNet. After volumetric partitioning, each block is fed into the backbone network (PointNet++~\cite{qi2017pointnetplusplus}) to extract point-wise features. Two loss functions, similarity loss and low-rank loss, are devised to predict the segmentation result for the block.}
  \label{fig:patchnet}\vspace{-8pt}
\end{figure*}

Given a 3D shape, the PriorNet is learned to delineate its meaningful parts (Figure~\ref{fig:patchnet}).
When the input is a partial shape (e.g., a partition block of the full shape),  the PriorNet would segment it
into patches which is either an independent part or a part of it. Ideally, PriorNet should avoid under-segmentation (i.e. grouping points belonging to two different parts) as much as possible.
PriorNet is trained to minimize a multi-task loss function defined for each block: $L=L_\text{sim}+L_\text{low-rank}$.

\paragraph{Volumetric partition and block re-sampling.}
Since it is difficult to directly use existing neural networks to process large point clouds, we perform a volumetric partition over the point clouds into blocks and then feed each block to the PriorNet.
%Meanwhile, this method can maximize to preserve the local details of block point sets.
In our implementation, the volume resolution is $7 \times 7 \times 7$.
After partitioning, empty blocks are removed. Since the number of points in each block varies a lot, to facilitate training, we re-sample the point set in each block into $D=512$ points, using farthest point sampling (FPS).

\paragraph{PriorNet architecture.}
In PriorNet, we use PointNet++~\cite{qi2017pointnetplusplus} as the backbone network to extract features for each point, using the default hyper-parameters in the original work.
Based on the point features, we define for each block
a similarity matrix to describe for any two points in the block how probable they belong to the same part.
This similarity matrix possesses low rank property with the rank
equal to the number of fine-grained parts in that block.
Therefore, part segmentation can be obtained by minimizing the rank of the matrix.

\paragraph{Similarity loss.}
For the purpose of point cloud segmentation, we aim to learn point features so that any two points belonging to the same part are as similar as possible, while those in different parts are as dissimilar as possible.
To this end, we defined a similarity matrix $S \in N_{d} \times N_{d}$ for the points in a black, where
$N_{d}$ is the number of points per block.
To estimate $S$, we design a similarity loss:
\begin{equation}
  L_\text{sim}=\sum_{i}^{N_d}\sum_{j}^{N_d}{l_{i,j}},
\label{eq:simloss}
\end{equation}
where
$$ l_{i,j}=\left\{
\begin{array}{lcl}
\|F(p_i)-F(p_j)\|_2,         \quad &   I(i,j)=1\\
\max\{0,K-\|F(p_i)-F(p_j)\|_2\}, \quad     & I(i,j)=0
\end{array} \right. $$
in which $I(i, j)$ indicates whether $p_i$ and $p_j$ belong to the same part in the ground-truth. $F$ is point feature extracted by PointNet++~\cite{qi2017pointnetplusplus}. $K$ controls the minimum dissimilarity between points in different parts; we use $K = 100$ by default.
We will show with experiment that our method is insensitive about the selection of $K$.

%During the inference phase, we use a threshold $b_{sim}$ to binarize the regressed similarity matrix. We set $b_{sim} = %100$ throughout our experiments.

\begin{figure}[t]
  \centering
  \includegraphics[width=3.3in]{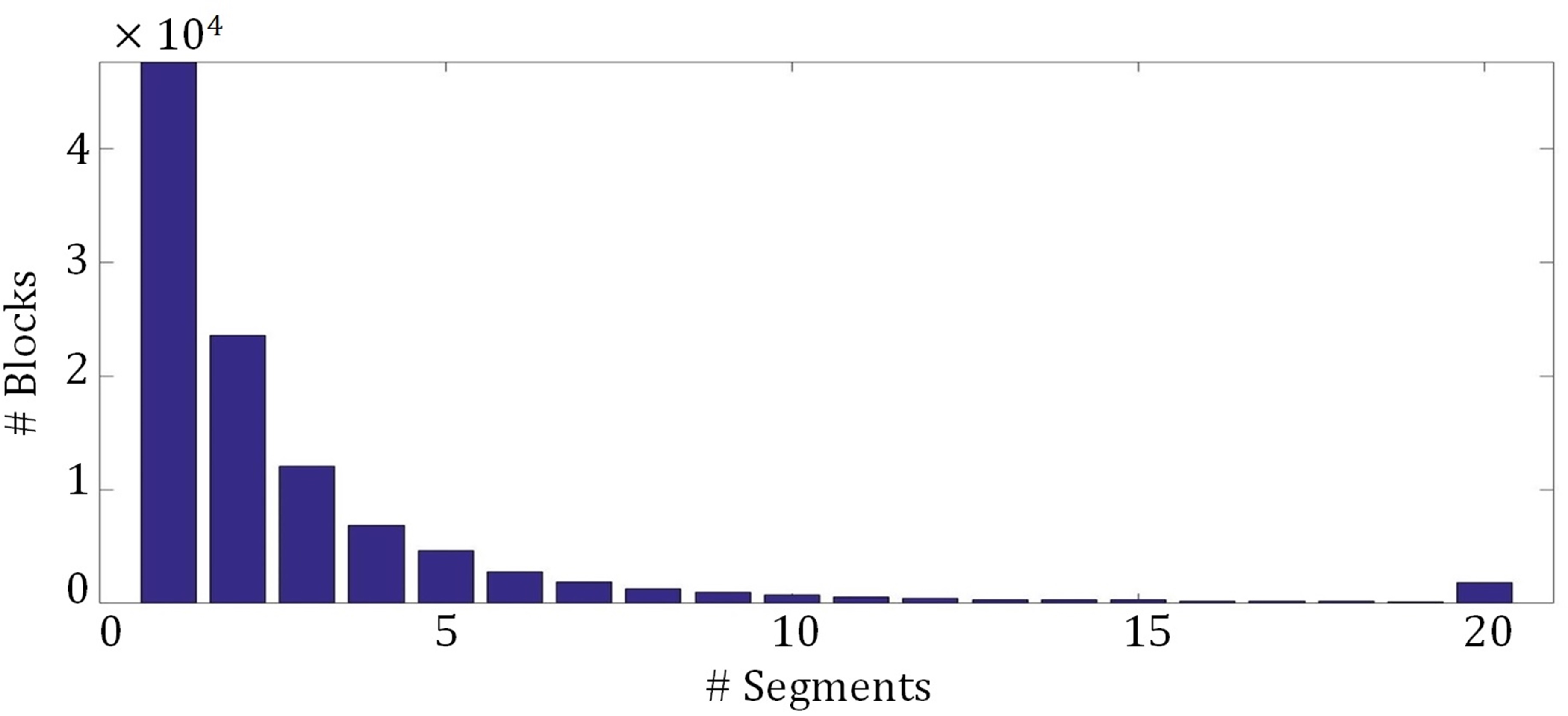}
  \caption{Statistics on segment count in a block over all the models in our dataset.}
  \label{fig:Segments_vs_blocks}
\end{figure}

\paragraph{Remarks on the low-rank property of the similarity matrix.}
The ``similarity'' here measures how probable two given points belong to the same part.
Ideally, the similarity matrix $S$ is defined
as an $N \times N$ matrix ($N$ is point count) with each entry $s_{i,j} = I(i,j)$ being an indicator
of whether point $p_i$ and $p_j$ belong to the same part.
It is obvious that the relation ``$p_i$ and $p_j$ belong to the same part'' is an equivalence relation.
This means that the corresponding relation graph has the following structure: the number of its connected components equals the number of parts, and each such connected component is a clique.
Clearly, the similarity matrix $S$ is the adjacency matrix of the relation graph.
If we rearrange the rows and columns of $S$ according to the parts, then $S$ is a
block-diagonal matrix in which each block consists of only ones, and zeros appear everywhere outside the blocks.
Each block corresponds to one part.
This block-diagonal matrix is obviously low-rank.

In reality, however, $S$ may not be a clean low rank matrix.
It can be ``contaminated'' due to noisy relations, becoming a probabilistic relation matrix with increased matrix rank.
Therefore, we design a low-rank loss that optimizes to recover the ideal similarity matrix and extract fine-grained segmentation result via minimizing the low-rank loss.

\paragraph{Low-rank approximation of similarity matrix.}
%After truncation by the parameter $K$, similarity matrix $S$ becomes a binarized matrix ($1$ indicates that two points %belongs to the same part, $0$ is not).

%We can observe that similarity matrix $S$ is a symmetric matrix, %(see Figure~\ref{fig:illumination_matrix_appro}),
%because $S(i,j)$ and $S(j,i)$ all represent the similarity between $p_i$ and $p_j$.

%because $S(i,j)$ and $S(j,i)$ all represent the simiarity for each pair of points ${p_i,p_j}$.
Ideally, the similarity matrix $S$ is a low rank matrix.
Let us denote $S = [S_1; S_2;\ldots; S_N]$, where $S_i$ represents $i$-th row of $S$.
If $p_i$ and $p_j$ belong to the same part, we have $S_i = S_j$.
%So similarity matrix $S$ should be a low rank matrix, and the rank $R$ equal to the number of parts.
%We assume that $S_x, S_y,..., S_r$ are all the different row vectors of the similarity matrix $S$, and we can easily %know that these vectors are orthogonal to each other.
Assuming that all linearly independent row vectors of $S$ are $\{S_{r_1}, S_{r_2},\ldots, S_{r_m}\}$, it is easy to verify that these row vectors are pair-wise orthogonal.
%Figure~\ref{fig:illumination_matrix_appro} shows an illustration of low-rank approximation.
%According to the low rank property property, we can conduct low rank decomposition on matrix $S$:
By using the elementary row transformation, $S$ can be represented by the maximal linear independent set of its row vectors:
\begin{equation}
\left\{
\begin{array}{ll}
  S_{1}=a_{1,1}S_{r_1} + a_{1,2}S_{r_2} + \ldots + a_{1,m}S_{r_m}\\
  S_{2}=a_{2,1}S_{r_1} + a_{2,2}S_{r_2} + \ldots + a_{2,m}S_{r_m}\\
  \cdots \\
  S_{N}=a_{N,1}S_{r_1} + a_{N,2}S_{r_2} + \ldots + a_{N,m}S_{r_m}\\
\end{array}\right.
\label{eq:eq1}
\end{equation}
Then we can rewrite (\ref{eq:eq1}) as
%\begin{equation}    \label{eq:eq2}
%S={[a_{1,1},a_{2,1},...,a_{N,1}]}^TS_{r_1} + \ldots +{[a_{1,m},a_{2,m},...,a_{N,m}]}^TS_{r_m}.   
%\end{equation}

\begin{eqnarray}    \label{eq:eq2}
S=&{[a_{1,1},a_{2,1},...,a_{N,1}]}^TS_{r_1} + \ldots    \\ 
 ~&+{[a_{1,m},a_{2,m},...,a_{N,m}]}^TS_{r_m}. \nonumber
\end{eqnarray}

If $p_i$ and $p_{r_1}$ belong to the same part, we know $S_i$ = $S_{r_1}$.
According to (\ref{eq:eq1}), we have $a_{i,r_1}=I(p_i,p_{r_1})$.
%\begin{equation}
%a_{i,r_1}=\left\{
%\begin{array}{ll}
%1,     &  \text{if $p_i$ and $p_{r_1}$ belong to the same part}\\
%0,     &  \text{otherwise}
%\end{array}\right.
%\label{eq:eq3}
%\end{equation}
Meanwhile, for each row vector $S_{r_1}$ = $[S_{1,{r_1}},S_{2,{r_1}},...,S_{N,{r_1}}]$,
we have $S_{i,r_1}=I(p_i,p_{r_1})$,
%\begin{equation}
%S_{i,{r_1}}=\left\{
%\begin{array}{ll}
%1,     &  \text{if $p_i$ and $p_{r_1}$ belong to the same part}\\
%0,     &  \text{otherwise}
%\end{array}\right.
%\label{eq:eq4}
%\end{equation}
where $S_{i,{r_1}}$ represents the $i$-th element of $S_{r_1}$.
Let us denote $A_{r_1} = [a_{1,1},a_{2,1},...,a_{N,1}]$, it holds that $A_{r_1} = S_{r_1}$,
%according to (\ref{eq:eq3})and(\ref{eq:eq4}), obviously,
%\begin{equation}
%   A_{r_1} = S_{r_1},
%\label{eq:eq5}
%\end{equation}
and similarly, $A_{r_2} = S_{r_2} \ldots, A_{r_m} = S_{r_m}$.
Substituting them into (\ref{eq:eq2}), we obtain

\begin{eqnarray}    \label{eq3}
S&=&A_{r_1}^TS_{r_1} + A_{r_2}^TS_{r_2} + ... + A_{r_m}^TS_{r_m}  \nonumber \\
~&=&S_{r_1}^TS_{r_1} + S_{r_2}^TS_{r_2} + ... + S_{r_m}^TS_{r_m}  \\
~&=&[S_{r_1}^T, S_{r_2}^T, ..., S_{r_m}^T]_{N \times R} {[S_{r_1}^T, S_{r_2}^T, ..., S_{r_m}^T]}_{N \times R}^T  \nonumber\\
~&=&M_{N \times R}  M_{N \times R}^T  \nonumber
\end{eqnarray}
which is a low rank decomposition of $S$, with $R$ being the rank of $S$.
%$M_{N \times R}$ represents the decomposed matrix.
%$M_{N_d \times R}^T$ is the transpose of $M_{N_d \times R}$.
%$N$ is the size of the point set. $R$ represents the rank of the similarity matrix $S$.
%$M_{N \times R}$ is also the segmentation mask of the point cloud, where each column represents a binary segmentation.

\paragraph{Low-rank approximation module.}
Since $S$ is symmetric, we only need to estimate $M_{N \times R}$.
To this end, we design a low-rank approximation module (Figure~\ref{fig:patchnet}).
In particular, we first use a three-layer MLP, with feature dimensions $512$, $256$, and $128$,
to transform the input similarity matrix features.
%Here, MLP($l_1,\ldots,l_d$) represents several multi-layer perceptrons (MLP) with output channels $l_1,\ldots,l_d$.
Since each point belongs to only one part, we then apply a softmax layer to assign each point feature a label in $(1,\ldots,R)$.
The output matrix is the predicted $M_{N \times R}$ with each column representing a part instance.

\paragraph{Low-rank loss.}
Based on the symmetric and low-rank properties of the similarity matrix, we design a low rank loss:
\begin{equation}
L_\text{low-rank} = ||M_{N \times R}  M_{N \times R}^T-S^\text{gt}||^2_2,
\end{equation}
where $M_{N \times R}$ is the predicted low rank matrix of the neural network. $S^\text{gt}$ is the ground-truth similarity matrix constructed using the training shapes.

Note that for each block, one cannot directly predict $M_{N \times R}$ because the actual rank $R$ is different from block to block.
According to statistics, we find that $98\%$ of the blocks has a segment count less than $5$. Therefore, we set the maximum rank to $5$ in our experiments.

In the training phase,
assuming that the segment count is $r$ in each block, we can extract the top $r$ columns from the predicted low rank matrix $M_{N \times R}$, obtaining $M_{N \times r}$.
Since each point belongs to only one part, we normalize the rows of $M_{N \times r}$.

In the testing phase, we make a prediction of the similarity matrix $S^\text{pred}$ and the low rank matrix $M_{N \times R}$.
To determine the segment count for each block,
%we cannot extract the top $r$ columns directly from the predicted low rank matrix $M_{N \times R}$ directly. In this case, we use a simple search to find the optimal $r$.
we first take the top $r$ columns ($r=\{1,2,\ldots,5\}$) to calculate the reconstructed similarity matrix $M_{N \times r}M_{N \times r}^T$ ($M_{N \times r}$ is row normalized), then select the $r$ which attains the minimum error between the reconstructed and the predicted similarity matrix.
%We select the lowest error $r$ as the predicted low-rank matrix.
%Finally, $M_{N \times r}$ with minimum error is selected as the estimated low-rank matrix.
The error is calculated as follows:
\begin{equation}
M_{N \times r} = \min_{1 \leq r \leq 5}{||M_{N \times r}  M_{N \times r}^T-S^\text{pred}||_2^2}.
\end{equation}
%where $M_{N \times r}$ is the top $r$ columns extracted from the predicted low rank matrix $M_{N \times R}$ and after $S^\text{pred}$ represents the predicted binary similarity matrix.

\paragraph{Training data preparation.}

\begin{figure}[t]
  \centering
  \includegraphics[width=2.9in]{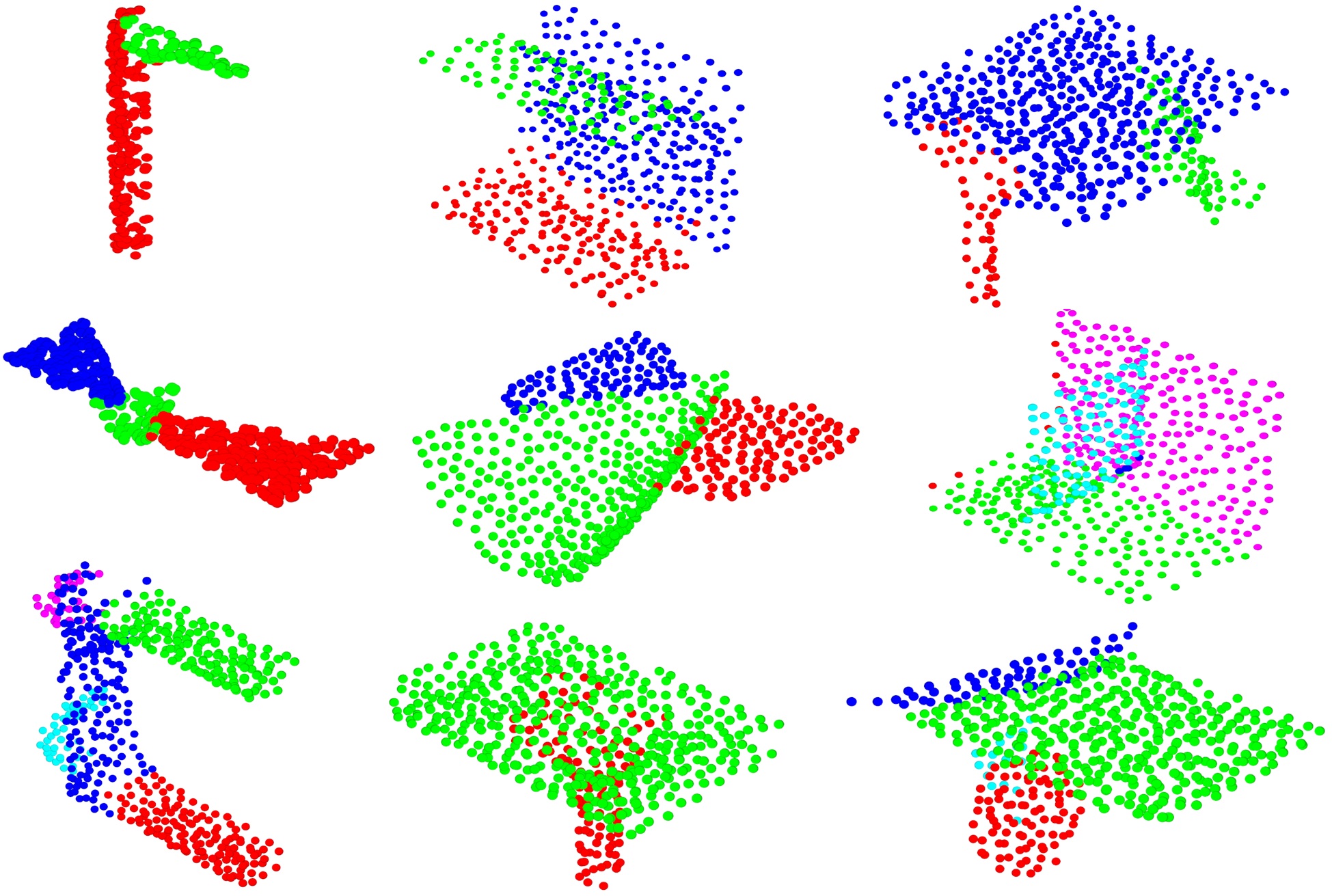}
  \caption{Some samples of block segmentation result.}
  \label{fig:blocks_segmentation}
\end{figure}

To train the PriorNet, we draw training blocks from $816$ training shapes.
Thanks to the block-wise training strategy, the required training shapes are greatly reduced.
Since both of the two training loss are related to the number of segments in blocks, we opt to balance training data for each segment count.
In particular, we randomly select 4K training blocks for each segment count as training samples, and the total number is 20K for all training blocks with segment count ranging from $1$ to $5$.
In Figure~\ref{fig:blocks_segmentation}, we show a few examples of block segmentation result.
PriorNet is quite effective in capturing the potential fine-grained parts in a block,
even for those with complicated structures.
Note that PriorNet is learned in a category-agnostic manner using blocks from all categories.

\subsection{Segment Merging Network (MergeNet)}
\label{sec:MergeNet}

\begin{figure*}[t]
  \centering
  \includegraphics[width=\textwidth]{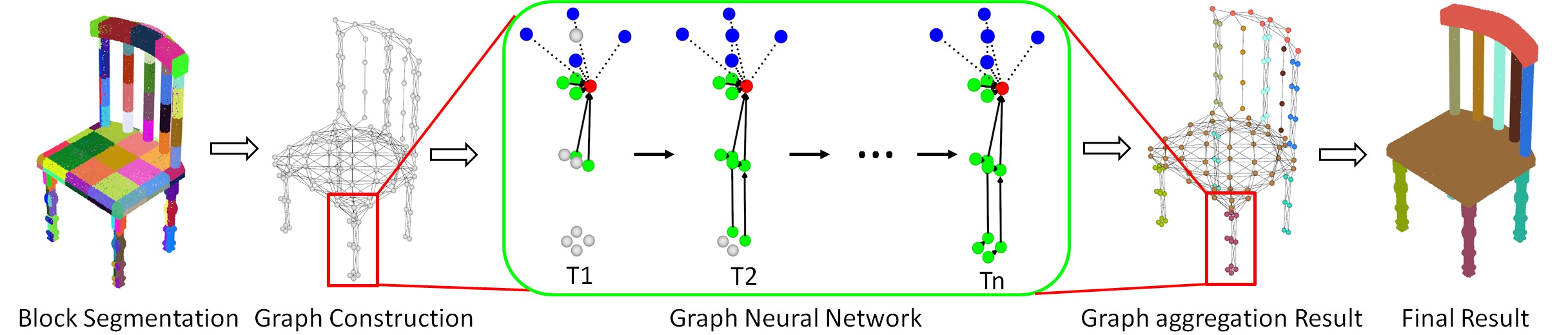}
  \caption{Network architecture of the MergeNet.
Given the segmentation results of all blocks, we construct a segment graph whose nodes are segments and edges represent the adjacency relation between segments. MergeNet is a graph convolution network trained over segment graphs, aiming to merge the segments of all blocks to form the final fine-grained segmentation.}
  \label{fig:mergenet}
\end{figure*}

Given the block segmentation results, the next step is to merge all the segments to form meaningful fine-grained parts for the whole shape.
To do so, we design a Graph Neural Network (GNN) which learns feature representation capturing not only local segment geometry but also global context.
A Graph Convolutional Network is a dynamic model in which the hidden representation of all nodes evolves over time. It can extract high-level node representation via message passing and produces a node-level output.

\paragraph{Segment graph construction.}
We first construct a graph $G = (\mathcal{V}, \mathcal{E}) $ whose nodes are the set of all segments and edges represent the adjacency relation between segments.
For each node, we compute its axis-aligned bounding box (AABB). Two nodes are neighboring if their AABB's intersect.
For each node $v \in \mathcal{V}$, we denote the input feature vector by $x_{v}$ and its hidden
representation describing the node’s state at network layer $l$ by $h_{v}^l$.
Let us use $\mathcal{N}_{v} $ to denote the set of neighboring nodes of $v$.
See Figure~\ref{fig:mergenet} for the visualization of a segment graph.

\paragraph{Graph propagation model.}
For each segment, we have extracted point-wise features with PriorNet in the previous stage. The features are fused together with max-pooling, serving as the initial hidden feature vector $x_v$ of segment $v$ to encode its local geometric information.
To make reliable decision in segment merging, contextual information is required, which can be obtained by the means of graph convolutions.

%Given that the graph already encodes the contextual information, this graph neural network exploits both local geometry and global context.
Each layer of the graph neural network can be written as a non-linear function:
\begin{eqnarray}
\begin{aligned}
  & a_{v}^{l} = \frac{1}{\mathcal{N}_{v}}\sum_{u \in \mathcal{N}_{v}}MLP({h_{u}^{l}}),\\
  & h_{v}^{l+1} = MLP([h_{v}^l, a_{v}^l]),
\label{eq:neural_layer}
\end{aligned}
\end{eqnarray}
where $a_{v}^{l}$ denotes the aggregation of the messages that node $v$ receives from its neighbors $\mathcal{N}_{v}$, for layer $l$. 
$MLP$ represents multi-layer perceptron.
When updating the hidden state, we first concatenate the hidden state $h_{v}^l$ and the message $ a_{v}^l $,
then we feed the concatenation to an MLP.
We use three layers of message passing in all our experiments.
%In this paper, we use vanilla RNN as the update function $F$ similar to~\cite{he2015delving}:
%\begin{eqnarray}
%\begin{aligned}
%  h_{s}^{l+1} = MLP([h_{s}^l, a_{s}^l]),
%\label{eq:update_fun}
%\end{aligned}
%\end{eqnarray}

\paragraph{Training loss.}
Similar to PriorNet, we also define a segment similarity matrix to encode whether two segments belong to the same fine-grained part.  For a similar reason, this similarity matrix also possesses low rank property.
For MergeNet, the parameters for the two loss functions (similarity and low rank) are the same as those for PriorNet except that the maximum rank is set to $100$.

\section{Results and Evaluations}
\label{sec:result}
%\todo{We present both qualitative and quantitative evaluations for each stage of our approach.}

%\subsection{Performance evaluation}
%\label{subsec:benchmark}

\paragraph{Implementation details.}
Our network is implemented with Tensorflow.
All training and testing shapes are densely sampled into 100K points to ensure that each tiny fine-grained parts are sufficiently sampled.
The resolution of the volumetric partition is $7 \times 7 \times 7$.
We use PointNet++~\cite{qi2017pointnetplusplus} as the backbone network  for feature extraction with the default hyperparameters in the original paper.
The batch size of PriorNet and MergeNet are $24$ and $4$, respectively.

\begin{figure}[t]
  \centering
  \includegraphics[width=2.9in]{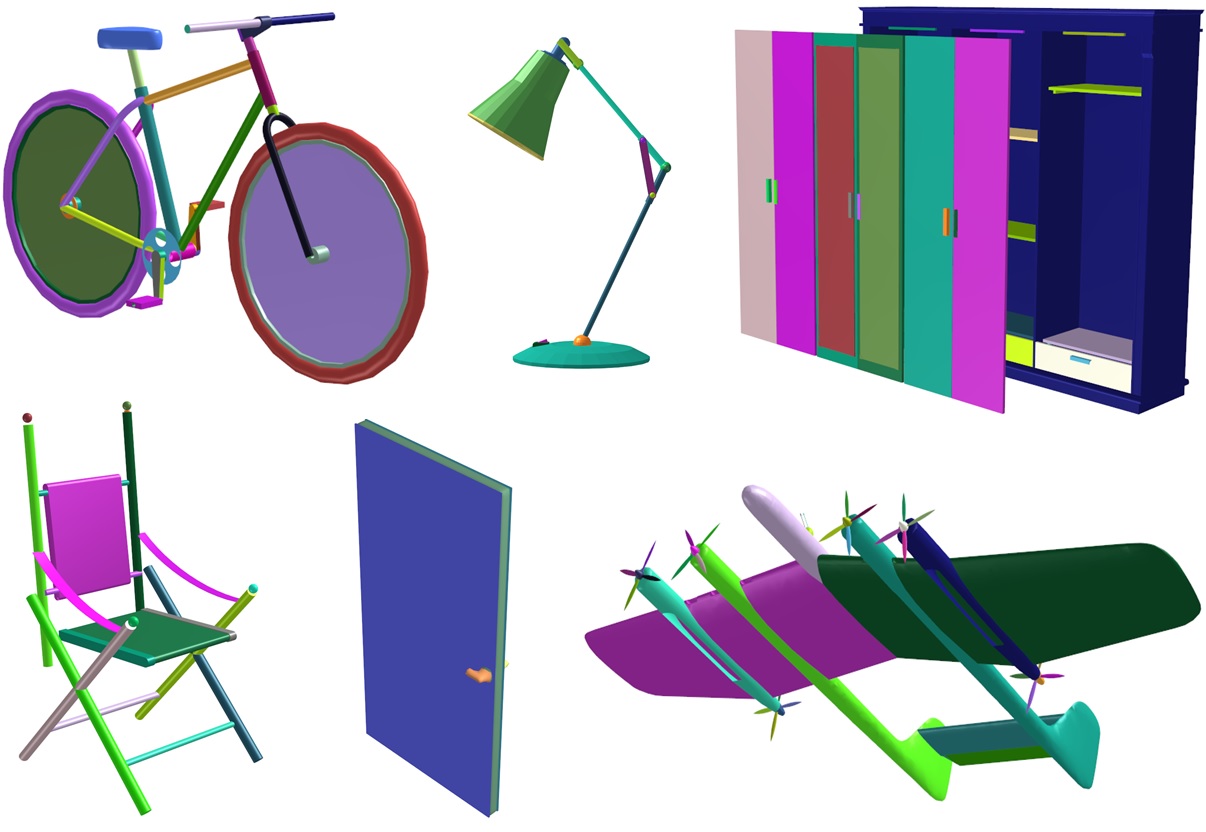}
  \caption{An overview of the fine-grained part dataset.}
  \label{fig:Dataset1}
\end{figure}

\begin{table*}[ht]
\caption{Accuracy of segmentation (average Intersection of Union, in percentage) on our dataset.
Row 1: The average number of fine-grained parts for each category.
Row 2: Training / testing split (number of models) of our dataset.
Row 3-4: Average IoU of PriorNet in different settings.
Row 5-8: Average IoU of SGPN~\cite{wang2018sgpn}, GSPN~\cite{yi2019gspn}, baseline, \wxg{our method(all parts) and our method (small parts)}.}
  \centering
    \begin{tabular*}{\textwidth}{lcccccccccc}
    \hline
    Rows  & Bed	& Chair	& Clock	& Door	& Lamp	& Table	& Cabinet	& Vehicle & Bicycle	 &Plane\\
    \hline\hline
    1. Avg. \#Parts        & 55.2	& 19.1	& 21.4	& 14.8	& 25.3	& 31.4	& 22.0	& 573.1	& 497.2	& 149.6\\
    2. \#Train  / \#Test  & 96/24 & 96/24  & 96/24  & 96/24 & 96/24 & 96/24  & 96/24  & 48/12 &  48/12
    &  48/12\\
    \hline
    \hline
    %{Patch-Net (ball partition)   } & 59.9 & 70.5	& 71.0	& 71.5 & 70.1 & 69.0 & 81.3	& 32.5 & 44.3 & 64.8\\
    3. {PriorNet (w/o low rank loss)} & 55.6 & 68.3	& 72.4	& 72.9 & 75.6 & 65.9 & 78.0	& 35.1 & 42.3 & 64.6\\
    4. {PriorNet                    } & 64.1 & 73.8	& 73.0  & 73.8 & 73.3 & 71.9 & 83.1	& 39.3 & 49.9 & 67.4\\
    \hline
    \hline
    5. SGPN~\cite{wang2018sgpn} & 19.1	& 39.8 & 23.8 & 34.0 & 33.4 & 40.7 &18.3  & 4.4	 & 6.1	& 11.4\\
    6. GSPN~\cite{yi2019gspn}         & 30.9	& 45.6 & 20.9 & 36.5 & 33.6 & 47.4 &25.3  & 5.8	 & 11.4	& 26.5\\
    7. Ours (PriorNet+BL)        & 30.7	& 42.4 & 38.7 & 38.7 & 35.1 & 44.9 &20.5  & 12.9	 & 15.8	& 20.6\\
    8. Ours (PriorNet+MergeNet) & \textbf{35.3} & \textbf{48.9} & \textbf{41.5} & \textbf{40.3} & \textbf{35.7} & \textbf{49.6} & \textbf{27.6} & \textbf{14.6} & \textbf{19.3} & \textbf{28.6}\\
    9. Ours (Small part)        & 19.4	& 37.9 & 24.1 & 36.5 & 34.1 & 46.3 & 24.6  & 12.5	 & 14.6	& 26.8\\
    \hline
  \end{tabular*}
\label{tab:LabellingAssessment}
\end{table*}

\paragraph{Training and testing data.}
To facilitate quantitative evaluation, we build a challenging dataset of 3D shapes
with highly fine-grained parts. These shapes were collected from the ShapeNet~\cite{ShapeNet2015} and the PartNet~\cite{mo2018partnet} datasets.  We choose $10$ commonly seen shape categories, including $7$ indoor categories and $3$ outdoor ones. The topological structure of the shapes within each category is quite diverse.
The second row of Table~\ref{tab:LabellingAssessment} reports the average number of parts per category.
Note that no part labels is available in our dataset.
%An overview of the benchmark dataset is given in Figure~\ref{fig:Dataset1}.
%This is mainly because it is very difficult to define the annotation for fine-grained parts. For fine-grained parts of the same %category models, it is also difficult to build a consistent fine-grained label when the topological structure is very different.

\paragraph{Timings.}
The training of PriorNet and MergeNet takes $17$ and $5.6$ hours for $100$ epochs on a NVIDIA TITIAN X GPU, respectively.
The testing time for each 3D point cloud is $6$ seconds for PriorNet and $4$ seconds for MergeNet.
The segment graph construction takes about $10$ seconds per shape.
The total computational time is about $20$ seconds per shape.

\begin{figure}[t]
  \centering
  \includegraphics[width=3.3in]{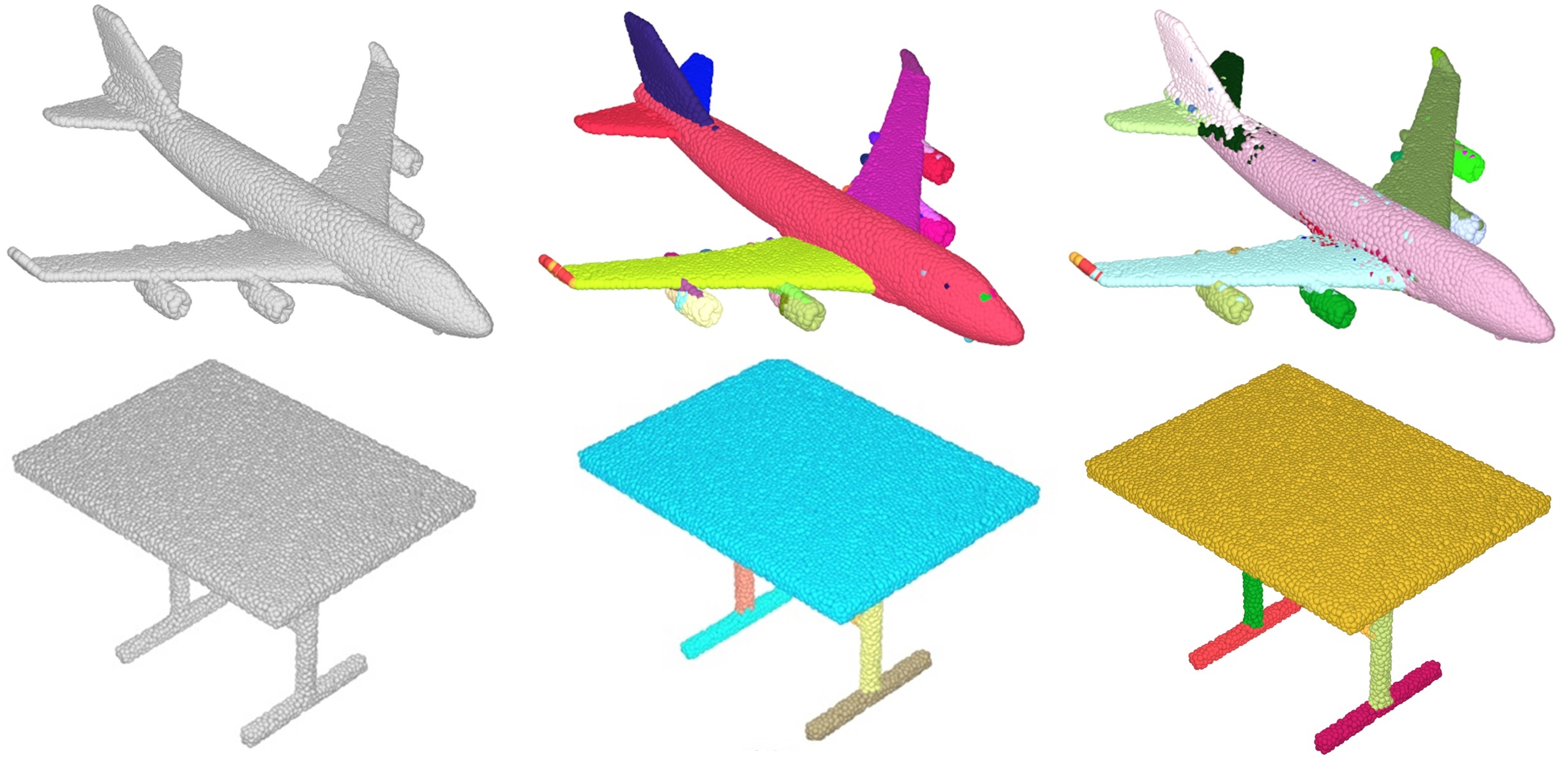}
  \caption{Some results of fine-grained segmentation.}
  \label{fig:more_result2}
\end{figure}

\begin{figure}[t]
  \centering
  \includegraphics[width=3.1in]{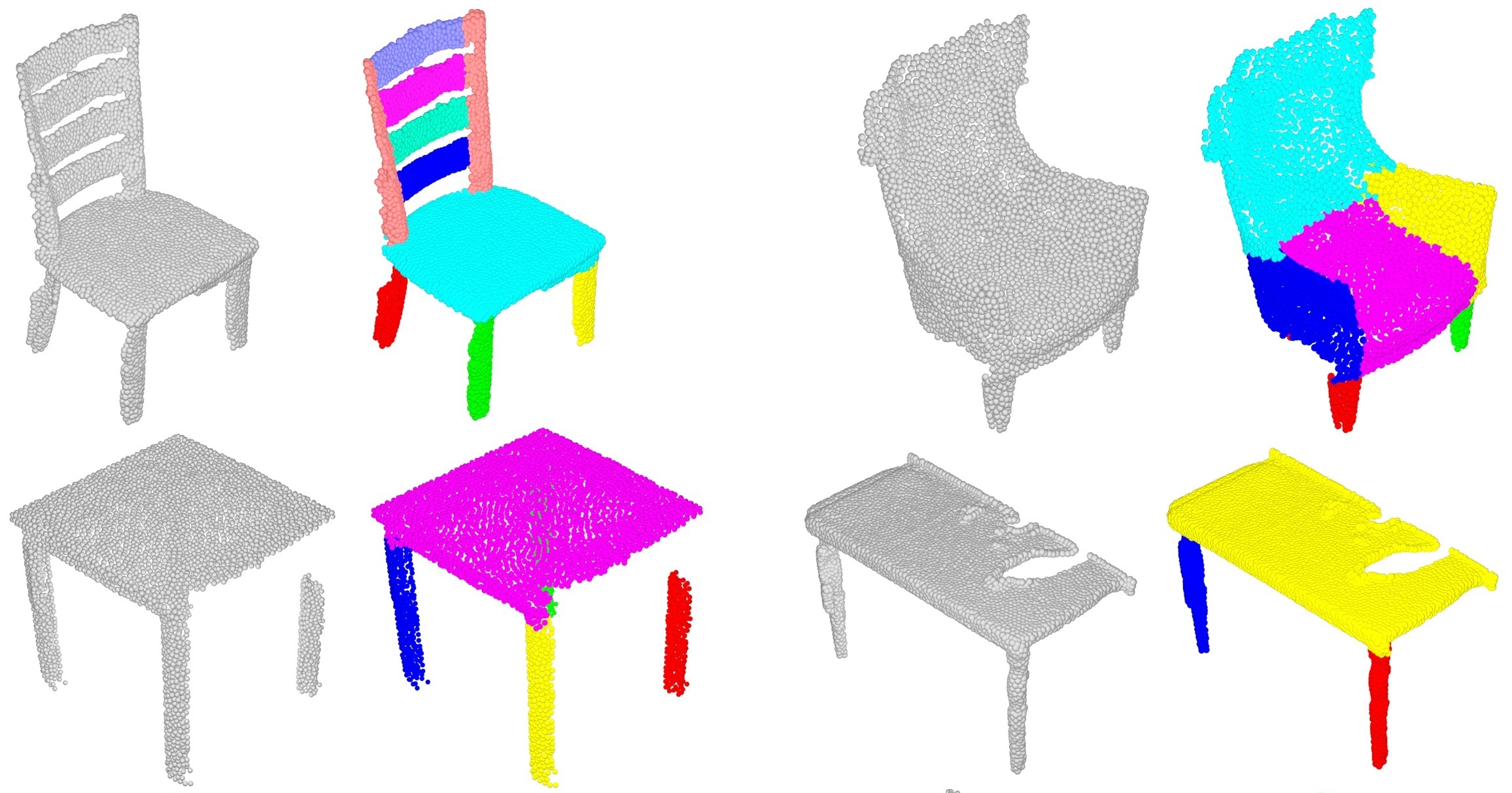}
  \caption{Segmentation results on real point cloud data by our method.}
  \label{fig:realdata}
\end{figure}

\paragraph{Quantitative and qualitative results.}
We train and test our model on our dataset, with a training/testing split of 8:2.
In our method, we train one PriorNet for all categories and one MergeNet per categoery.
The performance is measured by average Intersection of Union (avg. IoU) same as~\cite{Su_CVPR17}.
The results are reported in  row 8 of Table~\ref{tab:LabellingAssessment}.
\wxg{Meanwhile, in order to verify the effectiveness of our method for small part segmentation,
We also have independent statistics for these small parts.
The experimental results are reported in the loast row of table 1.
Note that in this paper, we cannot calculate the IOU directly since there is no part label.
Therefore, we design a simple strategy to calculate the IoU.
For each segmented fine-grained part, we first an IoU for each part of the GT segmentation. 
The maximum IoU is then taken as the part IoU. 
Finally, we take the average IoU over all segmented parts. 
This is a well-practiced scheme for estimating IoUs in segmentation without part tags (e.g. ~\cite{wang2018learning}).}
In Figure~\ref{fig:more_result2}, we show visually the fine-grained segmentation results.
We also test our method on real scan data (see Figure~\ref{fig:realdata}).

\paragraph{Comparison with the state of the arts.}
We compare our approach with SGPN ~\cite{wang2018sgpn} and GSPN ~\cite{yi2019gspn} , both of which are state-of-the-arts instance segmentation methods for 3D point clouds.
Their tasks are per-point labeling for segmentation.
To make a fair comparison, we made a simple modification to SGPN~\cite{wang2018sgpn} to fit our task,
which was to remove the semantic loss from the similarity matrix optimization.
For GSPN~\cite{yi2019gspn}, since our dataset does not contain semantic labels, we removed the semantic features from the feature backbone of R-PointNet and classification branch from the training tasks.

We report per-category IoU percentage on our dataset, see Row 5 and 6 of Table~\ref{tab:LabellingAssessment}. The results demonstrate the significant advantage of our method with more accurate segmentation.
There are two main reasons.
\emph{First}, since the number of fine-grained parts per model can be very large,
a holistic segmentation network must adopt a highly powerful backbone network to extract
robust and discriminant per-point feature targeting a large number of labels.
Our approach adopts a divide-and-conquer scheme to overcome this difficulty.
In our method, each block is segmented into a much smaller number of segments, which greatly reduces the difficulty in discriminant feature extraction.
\emph{Second}, the part count varies a lot and no part label is available in our dataset,
making it difficult to compute segmentation directly from the predicted similarity matrix~\cite{wang2018sgpn} or semantic
proposals~\cite{yi2019gspn}.
Our method achieves a robust segmentation prediction by optimizing the low rank loss.

\paragraph{Comparison with edge classification.}
To verify the effectiveness of our MergeNet, we design a baseline network which
classifies an edge of the segment graph to determine whether the involved two adjacent segments
belong to the same part.
%\supl{The network architecture and its detailed explanation can be found in the supplemental material.}
Taking a pair of adjacent segments as input, the network is trained to predict a score indicating whether the two segments belong to the same part.
We then employ a hierarchical aggregation algorithm to generate the final segmentation based on the predicted scores, similar to~\cite{wang_siga18}.
The results are shown in Table~\ref{tab:LabellingAssessment} (Row 7), which are inferior to those of our method.
The main reason is that the message passing process in the GCN aggregates global contextual information in the segment merging process, in contrast to the local prediction in the baseline method.

\begin{figure}[t]
  \centering
  \includegraphics[width=3.3in]{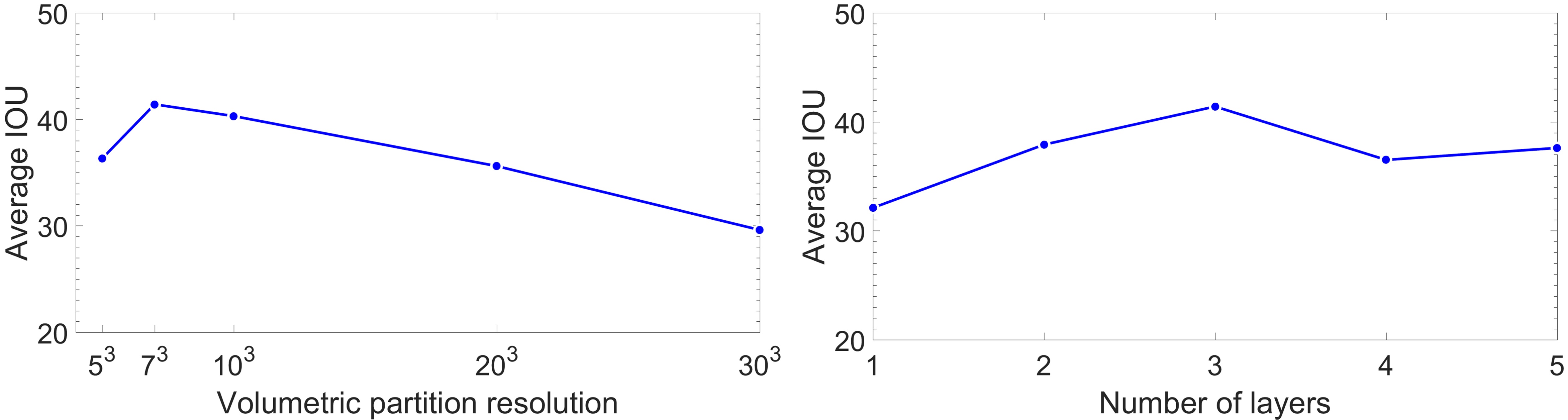}
  \caption{Segmentation accuracy (average IoU) over volumetric partition resolution (left) and the number of layers in the MergeNet (right).}
  \label{fig:voxel_layer_vs_iou}
\end{figure}

%\subsection{Parameters analysis}
%\label{subsec:patch_analysis}

\vspace{-1em}
\paragraph{Effect of low rank loss.}
To evaluate the effectiveness of our low rank loss, we experiment an ablated version of PriorNet which disables the low rank loss while keeping all other parameters unchanged. The experimental results are reported in Row 3 of
Table~\ref{tab:LabellingAssessment}. For all categories, our full method with low rank loss works better.
This is because the similarity matrix is usually noisy, the low rank constraint can be used to remove the noise effectively, thus improving the quality of segmentation.

\vspace{-1em}
\paragraph{Volumetric partition strategy.}
We evaluate the effect of the resolution of volumetric partition on segmentation quality.
%The voxel size $v$ is an important parameter of our method.
We experiment with the resolutions of $5^3$, $7^3$, $10^3$, $20^3$ and $30^3$, while keeping all other parameters the same. In Figure~\ref{fig:voxel_layer_vs_iou} (left), we plot the segmentation performance (Average IoU) over different resolution settings. The results are obtained by testing on all categories and taking the average.
The plot shows that best result is obtained at $7^3$.

\vspace{-1em}
\paragraph{Effect of layer count.}
To verify the design choice of the MergeNet, we experiment with different number of network layers.
Figure~\ref{fig:voxel_layer_vs_iou} (middle) shows the plot of average IoU over the number of layers.
The results are obtained by testing on all categories and taking the average.
The performance gradually improves as the number of layers increases, and then begins to oscillate.
We attribute this oscillation to overfitting.
We found that the best performance is obtained at $L = 3$.

\begin{figure}[t]
  \centering
  \includegraphics[width=3.2in]{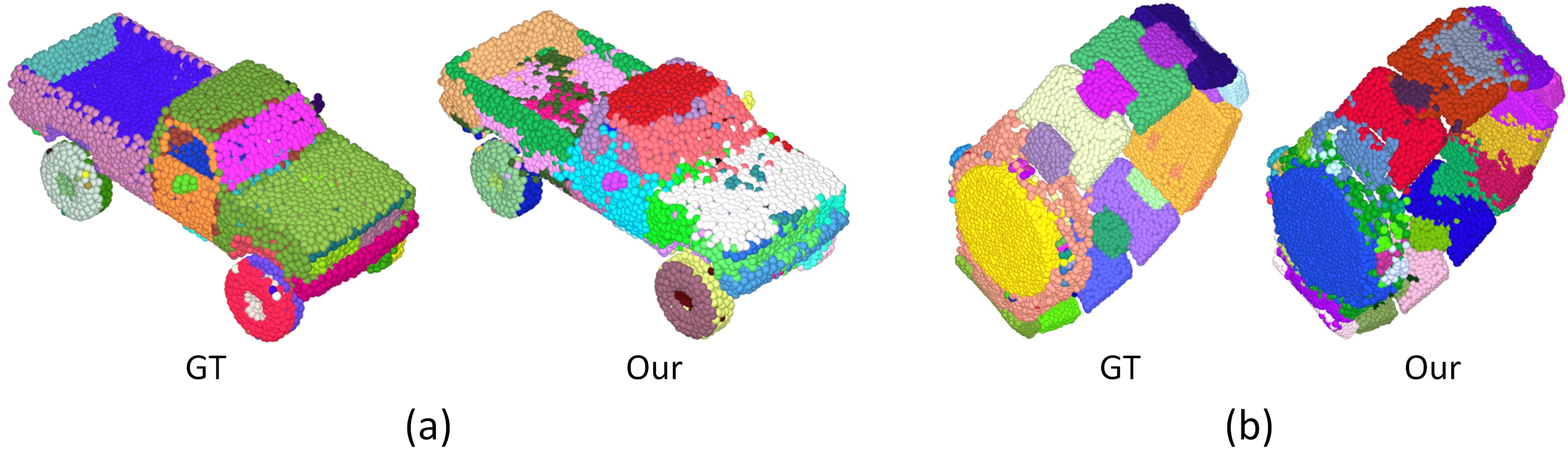}
  \caption{\wxg{Failure cases caused by setting a too large max-rank.}}
  \label{fig:failure_case}
\end{figure}

\begin{table}[!ht]
\caption{Comparison of AP between our method and two state-of-the-art fine-grained segmentation methods, i.e. PartNet~\cite{yu2019partnet} and SGPN~\cite{wang2018sgpn}.}
\centering
\setlength{\tabcolsep}{1.5mm}{
  \begin{tabular}{lcccccc}
    \hline
     Category & Aero & Bike & Chair & Helicopter & Sofa & Table \\			
    \hline\hline
    PartNet   & \textbf{88.4}  & \textbf{97.6} & \textbf{84.2}  & 69.4 & 55.8  & 63.2 \\
    SGPN              & 56.7  & 63.7 & 54.6  & 38.9 & 29.5  & 38.4\\
    Ours                    & 85.1  & 87.5 & 80.3  & \textbf{73.6} & \textbf{58.7}  & \textbf{71.6} \\
    \hline
  \end{tabular}
}
\label{tab:partnet}
\end{table}

\section{Conclusion}
\if 0
In this paper, we have presented a fine-grained segmentation algorithm for large scale
point sets. We approach the problem with a divide-and-conquer strategy.
At first, large point sets is partitioned as some blocks. For each block segmentation, we develop a network
Patch-Net. Block has three advantages:
1), Since the number of points in each block is relatively small, it can be easy to process large-scale point sets directly without sacrificing too many details.
2), The number of segments in each block is small, and it is easy to learn robust features to distinguish these segments in each block.
3), the key contribution of the network is a novel well-designed low rank loss.
Next, we design a graph convolutions based network, Merge-Net, to merge the segmentation results of adjacent blocks to obtain the fine-grained segmentation results.
\fi

We have presented a deep-clustering-based approach to the fine-grained segmentation of 3D point clouds.
The key idea is to learn geometric part priors which describe what constitutes a fine-grained part, from a dataset with fine-grained segmentation but no part semantic tags.
To handle large-scale point clouds, we adopt a divide-and-conquer scheme and split the input point cloud into a set of blocks. We train a deep neural network, PriorNet, to segment each block and then merge the segments into complete fine-grained parts using a graph neural network, MergeNet.

Our method has a few limitations:
\wxg{First, our method requires to set the parameter of maximum rank in low-rank approximation module.
However, a too large rank would cause noisy decomposition of the similarity matrix. 
Figure~\ref{fig:failure_case} shows two examples of the typical failure case could be inferior segmentation results caused by setting a too large max-rank.}
Second, our method is not designed to be end-to-end trainable due to the adoption of the divide-and-conquer scheme.
Third, the construction of segment graphs is time consuming.
In future, besides improving over the above limitations,
we would like to consider extending our method to handle 3D scene point clouds, exploiting
the advantage of our method in segmenting objects with significantly varying scales.

%Working with point sampled 3D shapes, our method models the clustering priors of 3D points with a similarity matrix of point features capturing how likely two points belong to the same part.
%This similarity matrix possesses low-rank property with the rank equal to the number of parts. Therefore, part segmentation can be obtained through minimizing a novel low rank loss.

%Furthermore, since fine-grained parts can be very tiny, a moderate sampling rate of shapes can hardly capture the geometry of tiny parts, which can result in suboptimal segmentation.
%To handle densely sampled point sets, we adopt a divide-and-conquer scheme.
%We first partition the large point set into a number of blocks.
%The similarity-based segmentation is performed over each block, based on a neural network, PriorNet, learned in
%a category-agnostic manner.
%We then train MergeNet, a graph convolution network, to merge the segments of all blocks to form the final segmentation result. 

\section*{Acknowledgement}
\wxg{We thank the anonymous reviewers for their valuable comments. This work was supported in part by 
National Key Research and Development Program of China (2018AAA0102200, 2019YFF0302902), and National Natural Science Foundation of China (61932003, 61532003).}

{\small
\bibliographystyle{ieee}
\bibliography{superpoints}
}

\end{document}